\documentclass[10pt,twocolumn,letterpaper]{article}

\newcommand{\real}{\mathbb{R}}
\newcommand{\myvec}[1]{\mathbf{#1}}
\newcommand{\myvecsym}[1]{\boldsymbol{#1}}

\newcommand{\vphi}{\myvecsym{\phi}}

\newcommand{\vPi}{\myvecsym{\Pi}}

\newcommand{\vh}{\myvec{h}}

\newcommand{\vw}{\myvec{w}}

\newcommand{\vx}{\myvec{x}}

\newcommand{\vy}{\myvec{y}}

\newcommand{\vB}{\myvec{B}}

\newcommand{\vD}{\myvec{D}}

\newcommand{\vG}{\myvec{G}}
\newcommand{\vH}{\myvec{H}}

\newcommand{\vK}{\myvec{K}}

\newcommand{\vP}{\myvec{P}}

\newcommand{\vS}{\myvec{S}}

\newcommand{\vU}{\myvec{U}}
\newcommand{\vV}{\myvec{V}}
\newcommand{\vW}{\myvec{W}}

\newcommand{\E}{\mathbb{E}}

\newcommand{\diag}{\mathrm{diag}}

\newcommand{\be}{\begin{equation}}
\newcommand{\ee}{\end{equation}}
\newcommand{\bea}{\begin{eqnarray}}
\newcommand{\eea}{\end{eqnarray}}
\newcommand{\beaa}{\begin{eqnarray*}}
\newcommand{\eeaa}{\end{eqnarray*}}

\DeclareMathAlphabet{\mathpzc}{OT1}{pzc}{m}{n}

\usepackage{iccv}
\usepackage{times}
\usepackage{epsfig}
\usepackage{graphicx}
\usepackage{url}
\usepackage{amsmath,amsthm,amssymb,amsfonts,amsthm}
\usepackage{wrapfig}
\usepackage{subfigure}
\usepackage{algorithm}
\usepackage{algorithmic}
\usepackage{mdwlist}
\usepackage{pbox}
\usepackage{fancyhdr}

\newcommand{\rbr}[1]{\left(#1\right)}

\newcommand{\cbr}[1]{\left\{#1\right\}}
\newcommand{\nbr}[1]{\left\|#1\right\|}

\newcommand{\RR}{\mathbb{R}}

\usepackage{hyperref}

\hypersetup{
  pagebackref=true,
  breaklinks=true,
  letterpaper=true,
  colorlinks,
  bookmarks=false
}

\iccvfinalcopy 

\def\iccvPaperID{1700} 

\ificcvfinal\fi
\begin{document}

\title{Deep Fried Convnets}

%

\author{
  Zichao Yang$^{1}$ \quad Marcin Moczulski$^{2}$ \quad Misha Denil$^{2,5}$\\
  {Nando de Freitas}$^{2,5,6}$ \quad {Alex Smola}$^{1,4}$ \quad {Le Song}$^3$
  \quad {Ziyu Wang}$^{2,5}$ \\[0.1cm]
  $^1$Carnegie Mellon University, $^2$University of Oxford, $^3$Georgia
  Institute of Technology \\ $^4$Google, $^5$Google DeepMind, $^6$Canadian Institute for Advanced Research \\
  \texttt{zichaoy@cs.cmu.edu, marcin.moczulski@stcatz.ox.ac.uk,}\\
  \texttt{mdenil@google.com, nandodefreitas@google.com,}\\
  \texttt{alex@smola.org, lsong@cc.gatech.edu, ziyu@google.com} }

\maketitle

\begin{abstract}
  The fully- connected layers of deep convolutional neural networks typically
  contain over 90\% of the network parameters. Reducing the number of
  parameters while preserving predictive performance is
  critically important for training big models in distributed systems and for deployment in embedded devices.

  In this paper, we introduce a novel Adaptive Fastfood transform to reparameterize the matrix-vector multiplication of fully connected layers. Reparameterizing a fully connected layer with $d$ inputs and $n$ outputs with the Adaptive Fastfood transform reduces the storage and computational costs costs from $\mathcal{O}(nd)$ to $\mathcal{O}(n)$  and $\mathcal{O}(n\log d)$ respectively. Using the Adaptive Fastfood transform in convolutional networks results in what we call a deep fried convnet. These convnets are end-to-end trainable, and enable us to attain substantial reductions in the number of parameters without affecting prediction accuracy on the MNIST and ImageNet datasets.
\end{abstract}

\section{Introduction}
\label{sec:intro}

In recent years we have witnessed an explosion of applications of convolutional neural networks with millions and billions of parameters. Reducing this vast number of parameters would improve the efficiency of training in distributed architectures. It would also allow for the deployment of state-of-the-art convolutional neural networks on embedded mobile applications. These train and test time considerations are both of great importance.

A standard convolutional network is composed of two types of layers, each with
very different properties.  Convolutional layers, which contain a small
fraction of the network parameters, represent most of the computational effort.
In contrast, fully connected layers contain the vast majority of the parameters
but are comparatively cheap to evaluate \cite{Krizhevsky:2014}.

This imbalance between memory and computation suggests that the efficiency of
these two types of layers should be addressed in different ways.
\cite{denton:2014} and \cite{Jaderberg:2014} both describe methods for
minimizing computational cost of evaluating a network at test time by approximating
the learned convolutional filters with separable approximations.  These
approaches realize speed gains at test time but do not address the issue of
training, since the approximations are made after the network has been fully
trained.  Additionally, neither approach achieves a substantial reduction in the number of parameters,
since they both work with approximations of the convolutional layers, which
represent only a small portion of the total number of parameters.  Many other works
have addressed the computational efficiency of convolutional networks in more
specialized settings \cite{farabet2010hardware, HongshengLi2014}.

In contrast to the above approaches,~\cite{DenilSDRF13} demonstrates that there
is significant redundancy in the parameterization of several deep learning
models, and exploits this to reduce the number of parameters. More
specifically, their method represents the parameter matrix as a product of two
low rank factors, and the training algorithm fixes one factor (called static
parameters) and only updates the other factor (called dynamic
parameters). \cite{SainathKSAR13} uses low-rank matrix factorization to reduce the size of the fully connected layers at train time. They demonstrate large improvements in reducing the number of parameters of the output softmax layer, but only modest improvements for the hidden fully connected layers. \cite{XueLG13} implements low-rank factorizations using the SVD after training the full model. In contrast, the methods advanced in \cite{SainathKSAR13} and this paper apply both at train and test time.  

In this paper we show how the number of parameters required to represent a deep
convolutional neural network can be substantially reduced without sacrificing
predictive performance.  Our approach works by replacing the fully connected
layers of the network with an Adaptive Fastfood transform, which is a generalization of the Fastfood transform for approximating kernels 
 \cite{le:2013}.  

Convolutional neural networks with Adaptive Fastfood transforms, which we refer to as deep fried convnets, are end-to-end trainable and
achieve the same predictive performance as standard convolutional networks on
ImageNet using approximately half the number of parameters.  

Several works have considered kernel methods in deep learning
\cite{huang:2014,ChoS09,song:2014,mairal:2014}. 
The Doubly Stochastic Gradients method of \cite{song:2014} showed that
effective use of randomization can allow kernel methods to scale to extremely
large data sets. However, the approach used fixed
convolutional features, and cannot jointly learn the kernel classifier and
convolutional filters. \cite{mairal:2014} showed how to learn a kernel function
in an unsupervised manner. 

There have been other attempts to replace the fully connected layers. The Network in Network architecture of
\cite{Lin:2014:ICLR} achieves state of the art results on several deep learning
benchmarks by replacing the fully connected layers with global average pooling.
A similar approach was used by \cite{Szegedy2014} to win the ILSVRC 2014 object
detection competition \cite{Russakovsky2014}.

Although the global average pooling approach achieves impressive results, it
has two significant drawbacks. First, feature transfer is more difficult with
this approach. It is very common in practice to take a convolutional network
trained on ImageNet and re-train the top layer on a different data set,
re-using the features learned from ImageNet for the new task (potentially with
fine-tuning), and this is difficult with global average pooling. This
deficiency is noted by \cite{Szegedy2014}, and motivates them to add an extra
linear layer to the top of their network to enable them to more easily adapt
and fine tune their network to other label sets. The second drawback of global
average pooling is computation.  Convolutional layers are much more expensive
to evaluate than fully connected layers, so replacing fully connected layers
with more convolutions can decrease model size but comes at the cost of
increased evaluation time. 

In parallel or after the first (technical report) version of this work, several researchers have attempted to create sparse networks by applying pruning or sparsity regularizers \cite{Collins2014,blundell-uncertainty-2015,Liu_2015_CVPR,Han2015}. These approaches however require training the original full model and, consequently, do not enjoy the efficient training time benefits of the techniques proposed in this paper. Since then,
hashing methods have also been advanced to reduce the number of parameters \cite{Chen2015,Bakhtiary15}. Hashes have irregular memory access patterns and, consequently, good performance on large GPU-based platforms is yet to be demonstrated.
Finally, distillation \cite{Hinton15,Romero2015} also offers a way of compressing neural networks, as a post-processing step.

\section{The Adaptive Fastfood Transform}
\label{sec:aft}

Large dense matrices are the main building block of fully connected neural network layers.  In propagating the signal from the $l$-th layer with $d$ activations $\vh_{l}$ to the $l+1$-th layer with $n$ activations $\vh_{l+1}$, we have to compute
\be
\vh_{l+1} = \vW \vh_{l}.
\ee
The storage and computational costs of this matrix multiplication step are both $\mathcal{O}(nd)$. The storage cost in particular can be prohibitive for many applications.

Our proposed solution is to reparameterize the matrix of parameters $\vW \in \mathbb{R}^{n \times d}$ with an Adaptive Fastfood transform, as follows
\be
\vh_{l+1} = \left(\vS\vH\vG\vPi \vH\vB\right) \vh_{l} = \widehat{\vW}\vh_{l}.
\ee

In Section~\ref{sec:intuitions}, we will provide background and intuitions behind this design. For now it suffices to state that the storage requirements of this reparameterization are $\mathcal{O}(n)$ and the computational cost is $\mathcal{O}(n \log d)$. We will also show in the experimental section that these theoretical savings are mirrored in practice by significant reductions in the number of parameters without increased prediction errors.

To understand these claims, we need to describe the component modules of the Adaptive Fastfood transform. For simplicity of presentation, let us first assume that $\vW \in \mathbb{R}^{d \times d}$. Adaptive Fastfood has three types of module:
\begin{itemize}
\item $\vS, \vG$ and  $\vB$ are diagonal matrices of parameters. In the original non-adaptive Fastfood formulation they are random matrices, as described further in Section~\ref{sec:intuitions}. The computational and storage costs are trivially $\mathcal{O}(d)$. 
\item $\vPi \in \{0,1\}^{d \times d}$ is a random permutation matrix. It can be implemented as a lookup table, so the storage and computational costs are also $\mathcal{O}(d)$.  
\item $\vH$ denotes the Walsh-Hadamard matrix, which is defined recursively as
$$\vH_2 := \begin{bmatrix} 1 & 1 \\ 1 & -1 \end{bmatrix} \mbox{  and }
\vH_{2d} := \begin{bmatrix} \vH_d & \vH_d \\ \vH_d & -\vH_d \end{bmatrix}.$$
The Fast Hadamard Transform, a variant of Fast Fourier Transform, enables us to
compute $\vH_d \vh_{l}$ in $\mathcal{O}(d \log d)$ time. 
\end{itemize}

In summary, the overall storage cost of the Adaptive Fastfood transform is 
$\mathcal{O}(d)$, while the computational cost is $\mathcal{O}(d \log d)$. These are substantial theoretical improvements over the $\mathcal{O}(d^2)$ costs of ordinary fully connected layers.

When the number of output units $n$ is larger than the number of inputs $d$, we can perform $n/d$ Adaptive Fastfood transforms and stack them to attain the desired size. In doing so, the computational and storage costs become $\mathcal{O}(n \log d)$ and $\mathcal{O}(n)$ respectively, as opposed to the more substantial $\mathcal{O}(nd)$ costs for linear modules. The number of outputs can also be refined with pruning.

\subsection{Learning Fastfood by backpropagation}
\label{sec:learning}

The parameters of the Adaptive Fastfood transform ($\vS, \vG$ and  $\vB$) can be learned by standard error derivative backpropagation. Moreover, the backward pass can also be computed efficiently using the Fast Hadamard Transform.

In particular, let us consider learning the $l$-th layer of the network,
$\vh_{l+1} = \vS\vH\vG\vPi \vH\vB \vh_{l}$.

For simplicity, let us again assume that 
$\vW \in \mathbb{R}^{d\times d}$ and that $\vh_{l} \in \RR^{d}$. Using backpropagation,
assume we already have $\frac{\partial E}{\partial \vh_{l+1}}$, where
$E$ is the objective function, then
\begin{equation}
  \frac{\partial E}{\partial \vS} = \diag \cbr{\frac{\partial E}{\partial \vh_{l+1}}
  (\vH \vG \vPi \vH \vB \vh_{l})^{\top}}.
\end{equation}
Since $\vS$ is a diagonal matrix, we only need to calculate the derivative with respect to 
the diagonal entries and this step requires only $\mathcal{O}(d)$ operations.

Proceeding in this way, denote the partial products by
\bea
\vh_{S} &=& \vH \vG \vPi \vH \vB \vh_{l} \nonumber \\
\vh_{H1} &=& \vG \vPi \vH \vB \vh_{l}\nonumber \\
\vh_{G} &=& \vPi \vH \vB \vh_{l}\nonumber \\
\vh_{\Pi} &=& \vH \vB \vh_{l}\nonumber \\
\vh_{H2} &=& \vB \vh_{l}. 
\eea
Then the gradients with respect to different parameters in the Fastfood layer can be computed recursively as follows:
\begin{align}
  \frac{\partial E}{\partial \vh_{S}} & = \vS^{\top}\frac{\partial
                                      E}{\partial \vh_{l+1}} &
  \frac{\partial E}{\partial \vh_{H1}} & = \vH^{\top}\frac{\partial E}{\partial
    \vh_{S}} \nonumber \\
  \frac{\partial E}{\partial \vG} & = \diag \cbr{\frac{\partial E}{\partial \vh_{H1}}
    \vh_{G}^{\top}} &
  \frac{\partial E}{\partial \vh_{G}} & = \vG^{\top}\frac{\partial E}{\partial
    \vh_{H1}} \nonumber \\
  \frac{\partial E}{\partial \vh_{\Pi}} & = \vPi^{\top} \frac{\partial
                                        E}{\partial \vh_{G}} &
  \frac{\partial E}{\partial \vh_{H2}} & = \vH^{\top}\frac{\partial E}{\partial
    \vh_{\Pi}} \nonumber \\
  \frac{\partial E}{\partial \vB} & = \diag \cbr{\frac{\partial E}{\partial
    \vh_{H2}} \vh_{l}^{\top} } &
  \frac{\partial E}{\partial \vh_{l}} & = \vB^{\top}\frac{\partial E}{\partial
    \vh_{H2}}.
\end{align}
Note that the operations in $\frac{\partial E}{\partial \vh_{H1}}$ and
$\frac{\partial E}{\partial \vh_{H2}}$ are simply applications of the Hadamard
transform, since $\vH^\top = \vH$, and consequently can be computed in $\mathcal{O}(d\log d)$
time. The operation in $\frac{\partial E}{\partial \vh_{\Pi}}$ is an application
of a permutation (the transpose of permutation matrix is a permutation matrix)
and can be computed in $\mathcal{O}(d)$ time. All other operations are diagonal matrix
multiplications.

\section{Intuitions behind Adaptive Fastfood}
\label{sec:intuitions}

The proposed Adaptive Fastfood transform may be understood either as a trainable type of structured random projection or as an approximation to the feature space of a learned kernel. Both views not only shed light on Adaptive Fastfood and competing techniques, but also open up room to innovate new techniques to reduce computation and memory in neural networks.    

\subsection{A view from structured random projections}

Adaptive Fastfood is based on the Fastfood transform \cite{le:2013}, in which the diagonal matrices $\vS$, $\vG$ and  $\vB$ have random entries. In the experiments, we will compare the performance of the existing random and proposed adaptive versions of Fastfood when used to replace fully connected layers in convolutional neural networks. 

The intriguing idea of constructing neural networks with random weights has been reasonably explored in the neural networks field \cite{Saxe2011,Jaeger02042004}. This idea is related to random projections, which have been deeply studied in theoretical computer science \cite{Mitzenmacher2005}. In a random projection, the basic operation is of the form 
\be
\vy = \vW \vx,
\ee
where $\vW$ is a random matrix, either Gaussian \cite{Indyk1998} or binary \cite{Achlioptas2003}. Importantly, the embeddings generated by these random projections approximately preserve metric information, as formalized by many variants of the celebrated Johnson-Lindenstrauss Lemma. 

The one shortcoming of random projections is that the cost of storing the matrix $\vW$ is $\mathcal{O}(nd)$.  Using a sparse random matrix $\vW$ by itself to reduce this cost is often not a viable option because the variance of the estimates of $\|\vW\vx\|$ can be very high for some inputs, for example when $\vx$ is also sparse. To see this, consider the extreme case of a very sparse input $\vx$, then many of the products with $\vW$ will be zero and hence not help improve the estimates of metric properties of the embedding space.

One popular option for reducing the storage and computational costs of random projections is to adopt random hash functions to replace the random matrix multiplication. For example, the count-sketch algorithm \cite{Charikar2004} uses pairwise independent hash functions to carry this job very effectively in many applications \cite{Cormode2012}. This technique is often referred to as the hashing trick \cite{Weinberger2009} in the machine learning literature. 
Hashes have irregular memory access patterns, so it is not clear how to get good performance on GPUs when following this approach, as pointed out in \cite{Chen2015}.

Ailon and Chazelle \cite{Ailon2009} introduced 
an alternative approach that is not only very efficient, but also preserves most of the desirable theoretical properties of random projections. Their idea was to replace the random matrix by a transform that mimics the properties of random matrices, but which can be stored efficiently. In particular, they proposed the following PHD transform:
\be
\vy = \vP \vH \vD \vx,
\ee
where $\vP$ is a sparse $n \times d$ random matrix with Gaussian entries, $\vH$ is a Hadamard matrix and $\vD$ is a diagonal matrix with $\{+1,-1\}$ entries drawn independently with probability $1/2$. The inclusion of the Hadamard transform avoids the problems of using a sparse random matrix by itself, but it is still efficient to compute.

We can think of the original Fastfood transform
\be
\vy = \vS\vH\vG\vPi \vH\vB \vx
\ee
as an alternative to this. Fastfood reduces the computation and storage of random projections to
$\mathcal{O}(n \log d)$ and $\mathcal{O}(n)$ respectively. In the original formulation $\vS, \vG$ and  $\vB$ are diagonal random matrices, which are computed once and then stored. 

In contrast, in our proposed Adaptive Fastfood transform, the diagonal matrices are learned by backpropagation. By adapting $\vB$, we are effectively implementing  Automatic Relevance
Determination on features. The matrix $\vG$ controls the bandwidth of the kernel and its spectral incoherence. Finally, $\vS$ represents different kernel types. For example, for the RBF kernel $\vS$ follows Chi-squared distribution. By adapting $\vS$, we learn the correct kernel type.

While we have introduced Fastfood in this section, it was originally proposed as a fast way of computing random features to approximate kernels. We expand on this perspective in the following section.

\subsection{A view from kernels}

There is a nice duality between inner products of features and kernels. This duality can be used to design neural network modules using kernels and vice-versa. 

For computational reasons, we often want to determine the features associated with a kernel. Working with features is preferable when the kernel matrix $\vK$ is dense and large. (Storing this matrix requires
$O(m^2)$ space, and computing it takes $O(m^2 d)$ operations, where $m$ is the
number of data points and $d$ is the dimension.) We might also want to design statistical methods using kernels and then map these designs to features that can be used as modules in neural networks. Unfortunately, one of the difficulties with this line of attack is that deriving features from kernels is far from trivial in general.

An important fact, noted in
\cite{Rahimi:2007}, is that infinite kernel expansions can be approximated
in an unbiased manner using randomly drawn features. For
shift-invariant kernels this relies on a classical result from harmonic
analysis, known as Bochner's Lemma, which states that 
  a continuous shift-invariant kernel $k(\vx,\vx')=k(\vx-\vx')$ on $\real^d$
  is positive definite if and only if $k$ is the Fourier transform of a
  non-negative measure $\mu(\vw)$.
This measure, known as the spectral density, in turn implies the existence of a
probability density $p(\vw)=\mu(\vw)/\alpha$ such that
\begin{align*}
    & k(\vx,\vx') =  \int \alpha e^{-i\vw^\top(\vx-\vx')} \,p(\vw) d\vw \\
    & = \alpha \E_\vw[\cos(\vw^\top\vx)\cos(\vw^\top\vx') 
      + \sin(\vw^\top\vx)\sin(\vw^\top\vx') ],
\end{align*}
where the imaginary part is dropped since both the kernel and distribution are
real.  

We can apply Monte Carlo methods to approximate the above expectation, and hence approximate the kernel $k(\vx,\vx')$ with an inner product of stacked cosine and sine features. 
Specifically,
suppose we sample $n$ vectors $i.i.d.$ from $p(\vw)$ and collect them
in a matrix $\vW=(\vw_1,\dots \vw_n)^\top$. The kernel can then be
approximated as the inner-product of the following random features:
\begin{align}
\vphi_\mathrm{rbf}(\vW\vx) = \sqrt{\alpha/n}\rbr{\cos(\vW\vx), \sin(\vW\vx)}^\top.
\label{eqn:feature}
\end{align}
That is, $\vphi(\vW\vx)$ is the neural network module, consisting of a linear layer $\vW\vx$ and entry-wise nonlinearities (cosine and sine in the above equation), that corresponds to a particular implicit kernel function.  

Approximating a given kernel function with random features requires the specification of a sampling distribution $p(\vw)$. Such distributions have
been derived for many popular kernels. For example, if we want the implicit kernel to be a squared exponential kernel,
\begin{align}
  k(\vx, \vx^{\prime}) = \exp\rbr{-\frac{\nbr{\vx-\vx'}^2}{2\ell^2}},
\end{align}
we know that the distribution $p(\vw)$ must be Gaussian: $\vw \sim \mathcal N(0, \diag(\boldsymbol\ell^2)^{-1})$. In other words, if we draw the rows of $\vW$ from this Gaussian distribution and use equation~(\ref{eqn:feature}) to implement a neural module, we are implicitly approximating a squared exponential kernel.

As another example of the mapping between kernels and random features, 
\cite{ChoS09,PandeyD14} introduced the rotationally invariant arc-cosine kernel
\be
  k(\vx, \vx^{\prime}) = \frac{1}{\pi}||\vx|| ||\vx'||
  (\sin(\theta) + (\pi - \theta) \cos(\theta)),
\ee
where $\theta$ is the angle between $\vx$ and $\vx^{\prime}$. Then by choosing $\vW$ to be a random Gaussian matrix, they showed that this kernel can be approximated with Rectified Linear Unit (ReLU) features:
\begin{align}
  \vphi_{\mathrm{relu}}(\vW \vx) & = \sqrt{1/n}\max(0, \vW \vx)
  \label{eq:relu-feature}.
\end{align}

The Fastfood transform was introduced to replace $\vW\vx$ in Equation~\ref{eqn:feature} with  $\vS\vH\vG\vPi \vH\vB\vx$, thus decreasing the computational and storage costs. 

\section{Deep Fried Convolutional Networks}
\label{sec:deepfried}

\begin{figure}[tbh]
  \includegraphics[width=0.98\linewidth]{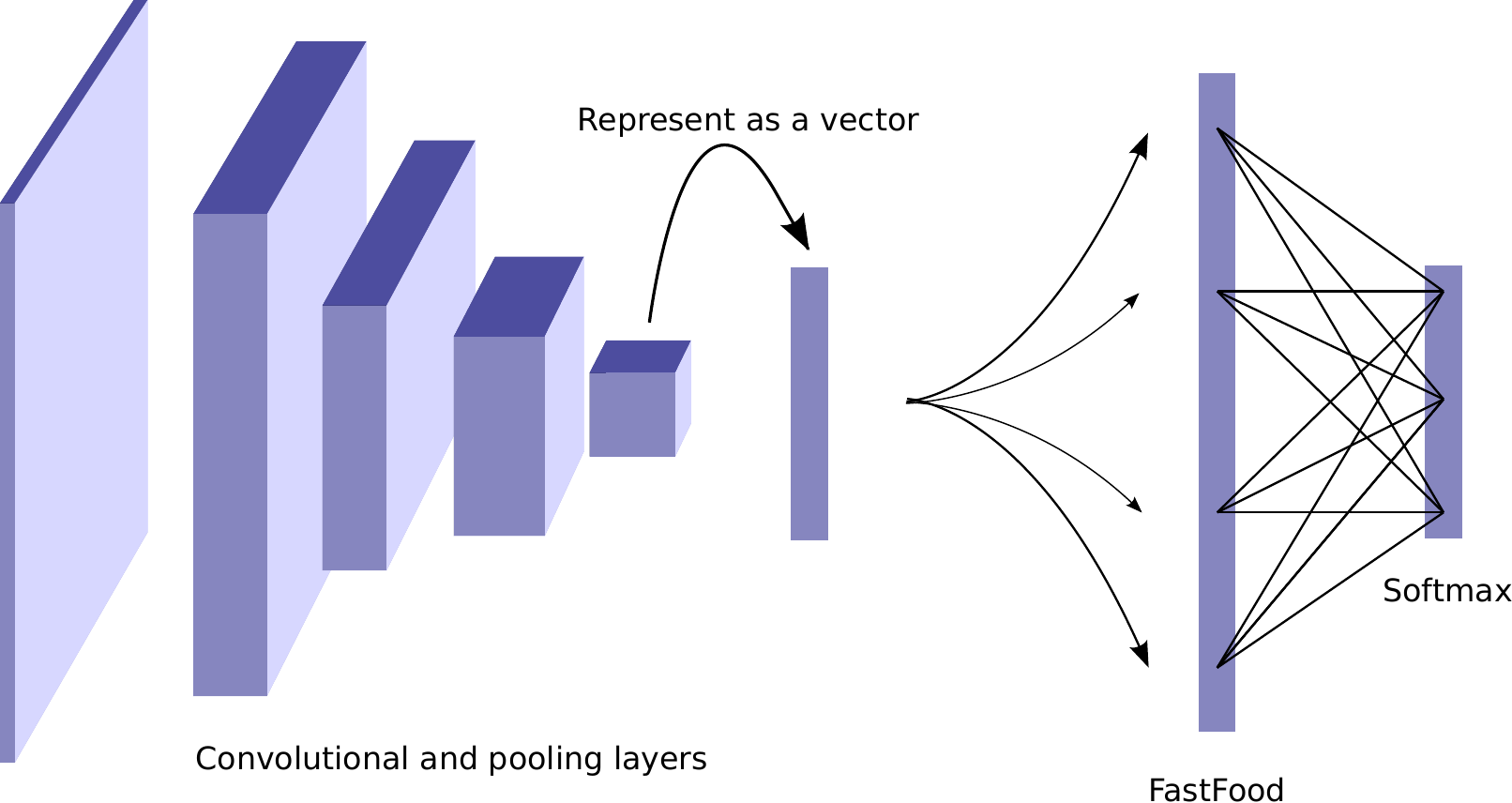}
  \centering
  \caption{The structure of a deep fried convolutional network.  The convolution and pooling layers are identical to those in a
    standard convnet. However, the fully connected layers are
    replaced with the Adaptive Fastfood transform.}
  \label{fig:ffconvnet}
\end{figure}


We propose to greatly reduce the number of parameters of the fully connected layers by replacing
them with an Adaptive Fastfood transform followed by a nonlinearity. We call this new
architecture a deep fried convolutional network. An
illustration of this architecture is shown in 
Figure~\ref{fig:ffconvnet}.

In principle, we could also apply the Adaptive Fastfood transform to the softmax classifier. However, reducing the memory cost of this layer is already well studied; for example, \cite{SainathKSAR13} show that low-rank matrix factorization can be applied during training to reduce the size of the softmax layer substantially. Importantly, they also show that training a low rank factorization for the internal layers performs poorly, which agrees with the results of \cite{DenilSDRF13}.  For this reason, we focus our attention on reducing the size of the internal layers.

\section{MNIST Experiment}

The first problem we study is the classical MNIST optical character recognition task.  This simple task serves as an easy proof of concept for our method, and contrasting the results in this section with our later experiments gives insights into the behavior of the Adaptive Fastfood transform at different scales.

As a reference model we use the Caffe implementation of the LeNet convolutional
network.\footnote{\url{https://github.com/BVLC/caffe/blob/master/examples/mnist/lenet.prototxt}}
It achieves an error rate of $0.87\%$ on the MNIST dataset.

We jointly train all layers of the deep fried network
(including convolutional layers) from scratch. We compare both the adaptive and non-adaptive Fastfood transforms using 1024 and 2048 features.  For the non-adaptive transforms we report the best performance achieved by varying the standard deviation of the random Gaussian matrix over the set $\cbr{0.001, 0.005, 0.01, 0.05}$, and for the adaptive variant we learn these parameters by backpropagation as described in Section~\ref{sec:learning}.

\begin{table}
  \centering
  \begin{tabular}{l|r|r}
    MNIST (joint) & Error & Params\\
    \hline
    Fastfood 1024 (ND) & 0.83\% & 38,821\\
    Adaptive Fastfood 1024 (ND) & 0.86\% &  38,821\\ 
    Fastfood 2048 (ND) & 0.90\% & 52,124\\
    Adaptive Fastfood 2048 (ND) & 0.92\% &  52,124\\ 
    \hline
    Fastfood 1024 & \textbf{0.71}\% &  38,821\\ 
    Adaptive Fastfood 1024 & 0.72\% &  38,821\\ 
    Fastfood 2048 & \textbf{0.71}\% &  52,124\\ 
    Adaptive Fastfood 2048 & 0.73\% &  52,124\\ 
    \hline
    Reference Model & \textbf{0.87}\% & 430,500
\end{tabular}
\caption{ MNIST jointly trained layers:
  comparison between a reference convolutional network with
  one fully connected layer (followed by a densely connected softmax layer) and two deep fried networks on the
  MNIST dataset.  Numbers indicate the number of features used in the Fastfood transform. The results tagged with \emph{(ND)} were obtained wtihout dropout.
}
\label{tab:mnist_joint}
\end{table}

The results of the MNIST experiment are shown in Table~\ref{tab:mnist_joint}.  Because the width of the deep fried network is substantially larger than the reference model, we also experimented with adding dropout in the model, which increased performance in the deep fried case.
Deep fried networks are able to obtain high accuracy using only a small fraction of of parameters of the original network
(11 times reduction in the best case).  Interestingly, we see no benefit from adaptation in this experiment, with the more powerful adaptive models performing equivalently or worse than their non-adaptive counterparts; however, this should be contrasted with the ImageNet results reported in the following sections.


\section{Imagenet Experiments}

We now examine how deep fried networks behave in a more realistic setting with
a much larger dataset and many more classes. Specifically, we use the ImageNet
ILSVRC-2012 dataset which has 1.2M training examples and 50K validation
examples distributed across 1000 classes.

We use the the Caffe ImageNet
model\footnote{\url{https://github.com/BVLC/caffe/tree/master/models/bvlc_reference_caffenet}}
as the reference model in these experiments \cite{jia2014caffe}.  This model
is a modified version of AlexNet \cite{KrizhevskySH12}, and achieves $42.6\%$
top-1 error on the ILSVRC-2012 validation set.  The initial layers of this
model are a cascade of convolution and pooling layers with interspersed
normalization.  The last several layers of the network take the form of an MLP
and follow a 9216--4096--4096--1000 architecture.  The final layer is a
logistic regression layer with 1000 output classes.  All layers of this network
use the ReLU nonlinearity, and dropout is used in the fully connected layers to
prevent overfitting.

There are total of 58,649,184 parameters in the reference model, of which
58,621,952 are in the fully connected layers and only 27,232 are in the
convolutional layers. The parameters of fully connected layer take up $99.9\%$
of the total number of parameters. We show that the Adaptive Fastfood transform can be used to substantially reduce the number of parameters in this model.

\begin{table}[th]
  \centering
  \begin{tabular}{l|r|r}
    ImageNet (fixed) & Error & Params\\
    \hline
    Dai et al.~\cite{song:2014} & 44.50\% & 163M \\
    Fastfood 16 & 50.09\% & 16.4M\\ 
    Fastfood 32 & 50.53\% & 32.8M\\ 
    Adaptive Fastfood 16 & 45.30\% & 16.4M\\
    Adaptive Fastfood 32 & \textbf{43.77\%} & 32.8M\\
    \hline
    MLP & \textbf{47.76\%} & 58.6M\\ 
  \end{tabular}
  \caption{Imagenet fixed convolutional layers:
    {MLP} indicates that we re-train
    9216--4096--4096--1000 MLP (as in the original network)
    with the convolutional weights pretrained and fixed. Our method is
    \emph{Fastfood 16} and \emph{Fastfood 32}, using 16,384 and
    32,768 Fastfood features respectively.
    \cite{song:2014} report results of max-voting of 10 transformations of
    the test set.}
  \label{tab:imagenet_fixed}
\end{table}

\subsection{Fixed feature extractor}

Previous work on applying kernel methods to ImageNet has focused on building models on features extracted from the convolutional layers of a pre-trained network~\cite{song:2014}.  This setting is less general than training a network from scratch but does mirror the common use case where a convolutional network is first trained on ImageNet and used as a feature extractor for a different task.  

In order to compare our Adaptive Fastfood transform directly to this previous work, we extract features from the final convolutional layer of a pre-trained reference model and train an Adaptive Fastfood transform classifier using these features.  Although the reference model uses two fully connected layers, we
investigate replacing these with only a single Fastfood transform. We experiment with two sizes for this transform: \emph{Fastfood 16} and \emph{Fastfood 32} using 16,384 and
    32,768 Fastfood features respectively. Since the Fastfood transform is a composite module, we can apply dropout between any of its layers. In the experiments reported here, we applied dropout after the $\vPi$ matrix and after the $\vS$ matrix. We also applied dropout to the last convolutional layer (that is, before the $\vB$ matrix).

We also train an MLP with the same structure as the top layers of the reference model for comparison.  In this setting it is important to compare against the re-trained MLP rather than the jointly trained reference model, as training on features extracted from fixed convolutional layers typically leads to lower performance than joint training~\cite{Yosinski:2014}.

The results of the fixed feature experiment are shown in Table~\ref{tab:imagenet_fixed}.  Following \cite{Yosinski:2014} and \cite{song:2014} we observe that training on
ImageNet activations produces significantly lower performance than of the
original, jointly trained network.  Nonetheless, deep fried networks are able to outperform both the re-trained MLP model as well as the results in~\cite{song:2014} while using fewer parameters.

In contrast with our MNIST experiment, here we find that the Adaptive Fastfood transform provides a significant performance boost over the non-adaptive version, improving top-1 performance by 4.5-6.5\%.

\subsection{Jointly trained model}

Finally, we train a deep fried network from scratch on ImageNet.  With 16,384 features in the Fastfood
layer we lose less than 0.3\% top-1 validation performance, but the number of
parameters in the network is reduced from 58.7M to 16.4M which corresponds to a
factor of 3.6x.  By further increasing the number of features to 32,768, we are
able to perform 0.6\% better than the reference model while using approximately
half as many parameters.  Results from this experiment are shown in Table~\ref{tab:imagenet_joint}.

\begin{table}[th]
  \centering
  \begin{tabular}{l|r|r}
    ImageNet (joint) & Error & Params \\
    \hline
    Fastfood 16 & 46.88\% & 16.4M\\
    Fastfood 32 & 46.63\% & 32.8M\\
    Adaptive Fastfood 16 & 42.90\% & 16.4M\\
    Adaptive Fastfood 32  & \textbf{41.93\%} & 32.8M\\
    \hline
    Reference Model & \textbf{42.59\%} & 58.7M\\ 
  \end{tabular}
  \caption{Imagenet jointly trained layers. Our method is
    \emph{Fastfood 16} and \emph{Fastfood 32}, using 16,384 and
    32,768 Fastfood features respectively.
    \emph{Reference Model} shows the accuracy of the jointly
    trained Caffe reference model. }
  \label{tab:imagenet_joint}
\end{table}

Nearly all of the parameters of the deep fried network reside in the final
softmax regression layer, which still uses a dense linear transformation, and
accounts for more than 99\% of the parameters of the network.  This is
a side effect of the large number of classes in ImageNet.  For a data set with
fewer classes the advantage of deep fried convolutional networks would be even
greater.  Moreover, as shown by \cite{DenilSDRF13,SainathKSAR13}, the last layer often
contains considerable redundancy.  We also note that any of the techniques from \cite{Collins2014,Chen2015} could be applied to the final layer of a deep fried
network to further reduce memory consumption at test time. We illustrate this with low-rank matrix factorization in the following section.

\section{Comparison with Post Processing}

In this section we provide a comparison to some existing works on reducing the number of parameters in a convolutional neural network.  The techniques we compare against here are \emph{post-processing} techniques, which start from a full trained model and attempt to compress it, whereas our method trains the compressed network from scratch.

Matrix factorization is the most common method for compressing neural networks, and has proven to be very effective.  Given
  the weight matrix of fully connected layers $\vW \in \mathbb{R}^{d\times n}$, we
  factorize it as
\begin{align*}
  \vW = \vU\vS\vV^{\top},
\end{align*}
where $\vU \in \mathbb{R}^{d \times d}$ and $\vV \in \mathbb{R}^{n \times n}$ and $\vS$ is a $d\times
n$ diagonal matrix. In order to reduce the parameters, we truncate all but the $k$ largest singular values, leading to the approximation
$
  \vW \approx \tilde{\vU}\tilde{\vV}^{\top},
$
where $\tilde{\vU} \in \mathbb{R}^{d \times k}$ and $\tilde{\vV} \in \mathbb{R}^{n \times k}$ and $\vS$ has been absorbed into the other two factors.  If $k$ is sufficiently small then storing $\tilde{\vU}$ and $\tilde{\vV}$ is less expensive than storing $\vW$ directly, and this parameterization is still learnable.

It has been shown that training a factorized representation directly leads to poor performance~\cite{DenilSDRF13} (although it does work when applied only to the final logistic regression layer~\cite{SainathKSAR13}).  However, first training a full model, then preforming an SVD of the weight matrices followed by a fine tuning phase preserves much of the performance of the original model~\cite{XueLG13}.  We compare our deep fried approach to SVD followed by fine tuning, and show that our approach achieves better performance per parameter in spite of training a compressed parameterization from scratch.  We also compare against a post-processed version of our model, where we train a deep fried convnet and then apply SVD plus fine-tuning to the final softmax layer, which further reduces the number of parameters.

Results of these post-processing experiments are shown in Table~\ref{tab:imagenet_svd}. For the SVD decomposition of each of the three fully connected
layers in the reference model we set $k = \min(d, n)/2$ in SVD-half and
$k = \min(d, n)/4$ in SVD-quarter. SVD-half-F and SVD-quarter-F mean that the
model has been fine tuned after the decomposition.

There is 1\% drop in accuracy for SVD-half and 3.5\%
drop for SVD-quarter. Even though the increase in the error for the SVD can be mitigated by finetuning (the drop decreases to 0.1\% for SVD-half-F and 1.3\% for
SVD-quarter-F), deep fried convnets still perform better both in terms of the accuracy and the number of parameters.

Applying a rank 600 SVD followed by fine tuning to the final softmax layer of the Adaptive Fastfood 32 model removes an additional 12.5M parameters at the expense of $\sim$0.7\% top-1 error.

For reference, we also include the results of Collins and Kohli~\cite{Collins2014}, who pre-train a full network and use a sparsity regularizer during fine-tuning to encourage connections in the fully connected layers to be zero.  They are able to achieve a significant reduction in the number of parameters this way, however the performance of their compressed network suffers when compared to the reference model.  Another drawback of this method is that using sparse weight matrices requires additional overhead to store the indexes of the non-zero values.  The index storage takes up space and using sparse representation is better than using a dense matrix
  only when number of nonzero entries is small.

\begin{table}
  \centering
  \begin{tabular}{l|r|r|r}
    Model & Error & Params & Ratio \\
    \hline
    Collins and Kohli~\cite{Collins2014} & 44.40\% & --- & ---\\
    SVD-half & 43.61\% & 46.6M & 0.8 \\
    SVD-half-F & 42.73\% &  46.6M & 0.8 \\
    Adaptive Fastfood 32 & \textbf{41.93\%} & 32.8M & 0.55\\
    SVD-quarter&  46.12\% & 23.4M  & 0.5 \\
    SVD-quarter-F & 43.81\%  & 23.4M & 0.5 \\
    Adaptive Fastfood 16 & 42.90\% & 16.4M & 0.28 \\
    \hline
    Ada. Fastfood 32 (F-600) & 42.61\% & 20.3M & 0.35 \\
    \hline
    Reference Model & 42.59\% & 58.7M & 1\\
  \end{tabular}
  \caption{Comparison with other methods.
    The result of \cite{Collins2014} is based on
    the the Caffe AlexNet model (similar but not identical to the Caffe reference
    model) and achieves $\sim$4x reduction in memory usage,
    (slightly better than Fastfood 16 but with a noted drop in performance).
    SVD-half: 9216-2048-4096-2048-4096-500-1000 structure.
    SVD-quarter: 9216-1024-4096-1024-4096-250-1000 structure.
    \texttt{F} means after fine tuning.}
\label{tab:imagenet_svd}
\end{table}

\section{Conclusion}

Many methods have been advanced to reduce the size of convolutional networks at test time. In contrast to this trend, the Adaptive Fastfood transform introduced in this paper is end-to-end differentiable and hence it enables us to attain reductions in the number of parameters even at train time.

Deep fried convnets capitalize on the proposed Adaptive Fastfood transform to achieve a substantial reduction in the number of parameters without sacrificing predictive performance on MNIST and ImageNet. They also compare favorably against simple test-time low-rank matrix factorization schemes.

Our experiments have also cast some light on the issue of random versus adaptive weights. The structured random transformations developed in the kernel literature perform very well on MNIST without any learning; however, when moving to ImageNet, the benefit of adaptation becomes clear, as it allows us to achieve substantially better performance.  This is an important point which illustrates the importance of learning which would not have been visible from experiments only on small data sets.

The Fastfood transform allows for a theoretical reduction in computation from $\mathcal{O}(nd)$ to $\mathcal{O}(n\log d)$. However, the computation in convolutional neural networks is dominated by the convolutions, and hence deep fried convnets are not necessarily faster in practice.

It is clear looking at out results on ImageNet in Table~2 that the remaining parameters are mostly in the output softmax layer. The comparative experiment in Section~7 showed that the matrix of parameters in the softmax can be easily compressed using the SVD, but many other methods could be used to achieve this. One avenue for future research involves replacing the softmax matrix, at train and test times, using the abundant set of techniques that have been proposed to solve this problem, including low-rank decomposition, Adaptive Fastfood, and pruning.

The development of GPU optimized Fastfood transforms that can be used to replace linear layers in arbitrary neural models would also be of great value to the entire research community given the ubiquity of fully connected layers layers.


{\small
\bibliographystyle{ieee}
\bibliography{deepBib,embedding}
}


\end{document}